\documentclass[10pt, a4paper, oneside]{article}
\usepackage[hidelinks]{hyperref}
\usepackage{graphicx}
\usepackage{url}
\usepackage{ulem}
\usepackage{mathtools}
\usepackage{scalerel}
\usepackage{setspace}
\usepackage[strict]{changepage}
\usepackage{caption}
\usepackage[letterspace=-50]{microtype}
\usepackage{afterpage}
\usepackage{multirow}
\usepackage{float}
\usepackage{lscape}

\usepackage{titlesec}

\titleformat*{\section}{\Large\bfseries}
\titleformat*{\subsection}{\normalsize\bfseries}

\graphicspath{{./figures/}}

\usepackage[textwidth=8cm, margin=0cm, left=4.6cm, right=4.2cm, top=3.9cm, bottom=6.8cm, a4paper, headheight=0.5cm, headsep=0.5cm]{geometry}
\usepackage{fancyhdr}
\usepackage[format=plain, labelfont=it, textfont=it, justification=centering]{caption}
\usepackage{breakcites}
\usepackage{microtype}
 
\begin{document}

\title{Myers-Briggs personality classification from social media text using  pre-trained language models}


\author{Vitor Garcia dos Santos * \and Ivandr\'e Paraboni \thanks{Universidade de S\~ao Paulo, Brazil. E-mail: \{vitor.garcia.santos,ivandre\}@usp.br} }



\maketitle

{\fontfamily{ptm}\selectfont
\begin{abstract}
{\fontsize{9pt}{9pt}\selectfont{\vspace*{-2mm}
In Natural Language Processing, the use of pre-trained language models has been shown to obtain state-of-the-art results in many downstream tasks such as sentiment analysis, author identification and others. In this work, we address the use of these methods for personality classification from text. Focusing on the Myers-Briggs (MBTI) personality model, we describe a series of experiments in which the well-known Bidirectional Encoder Representations from Transformers (BERT) model is fine-tuned to perform MBTI classification. Our main findings suggest that the current approach significantly outperforms well-known text classification models based on bag-of-words and static word embeddings alike across multiple  evaluation scenarios, and generally outperforms previous work in the field.}}
\end{abstract}}

\section{Introduction}
\label{sec.intro}

Human personality  - a  set of relatively stable behaviour patterns of an individual \cite{allport} - has been the focus of studies in multiple disciplines, and it is   well-known to computer science through personality models such as the  Big Five \cite{b5-goldberg} and, perhaps to a lesser extent, the Myers-Briggs Type Indicator (MBTI) \cite{mbti}. Models of this kind associate word choices made by an individual (e.g., a  customer, a social media user etc.) to pre-defined personality categories (e.g., extroverts versus introverts), allowing us to assess their personality traits for a wide range of practical applications in both natural language interpretation \cite{ranlp-wesley,depress-lrec} and generation \cite{stars-mutual}.   

Personality assessment  may however  require the use of personality inventories (e.g., \cite{b5-tech}) with the aid of specialists, which may become costly in large scale. As an alternative to this,  studies in Natural Language Processing (NLP) and related fields have addressed the relation between language use and personality to develop methods for automatically detecting the personality traits of individuals based on text samples that they have written (e.g., on social media etc.) \cite{mbti-plank,personality2017,b5-mult,mbti-author}. 

Personality detection from text  may be seen as an instance of author profiling, that is, the task of inferring an author's demographics based on text samples that they have authored \cite{ca-fineg,ca-bert,ca-celeb,ca-richer,b5-ieee,b5-newrev,exp-delmondes}. As in other  profiling tasks (e.g., author's gender or age detection), studies of this kind are  usually implemented with the aid of supervised machine learning based on text corpora labelled with personality information. The issue of personality classification from text with a specific focus on the MBTI personality model is the subject of the present study.

Existing work in MBTI classification from text  generally follows much of the same methods seen elsewhere in NLP, which usually comprise the use of a bag-of-words model or, more recently, static word embeddings such as those provided by Word2vec \cite{word2vec} and similar approaches. We notice, however, that more recent text representation models - in particular, context-sensitive embeddings such as those provided by Bidirectional Encoder Representations from Transformers (BERT)  \cite{bert} - are still relatively uncommon in MBTI classification, even though these models have been shown to obtain state-of-the-art results in a wide range of NLP applications from  sentiment analysis \cite{as-bert} to author identification \cite{aa-pretrain}, and many others.

Based on these observations, the present work addresses the use of pre-trained BERT  language models for MBTI personality classification from text written in multiple  languages. In doing so, our  objective is to show that by fine-tuning BERT to the present task we may significantly outperform the use of other text representation models  across these evaluation scenarios, with two main contributions to the field: 

\begin{quote}
    \item[1]{BERT-based models for MBTI personality classification from text in multiple languages.}
    \item[2]{Robust, cross-validation results shown to be consistently superior to those obtained by bag-of-words and static word embeddings alike, and to previous work in the field.}
\end{quote}

The reminder of this paper is structured as follows. Section \ref{sec.background} reviews recent approaches to MBTI personality classification from text. Section \ref{sec.method} introduces a number of computational models for the task -  including the use of pre-trained language models and baseline alternatives -   and the datasets to be taken as a basis for our experiments. Section \ref{sec.results} presents results obtained by these models, and Section \ref{sec.final} summarises our findings and describes opportunities for future work.

\section{Background}
\label{sec.background}

In what follows we review existing work in MBTI personality classification, and briefly discuss opportunities for using pre-trained language models in this task.

\subsection{Related work}

Table \ref{tab.related} summarises a number of recent studies in MBTI personality classification from text   by  reporting the target language (Ar=Arabic, En=English, De=German, Du=Dutch, It=Italian, Fr=French, Pt=Portuguese, Sp=Spanish, In=Indonesian), domain (T=Twitter, R=Reddit, F=Facebook, O=online forums, E=essays, V=vlogs), machine learning method (lr=logistic regression, svm=support vector machine, nb=Naive Bayes, rf=Random Forest, ens=ensemble, seq=BERT sequence learner, svd=singular value decomposition, xg=XGBoost), and learning features (w=word, c=character, pos=part-of-speech, u=user attributes, n=network attributes, p=psycholinguistic features from LIWC \cite{liwc} and MRC \cite{mrc}, t=LDA topics \cite{lda}, w2v=Word2vec \cite{word2vec} and BERT \cite{bert} word embeddings, s=text statistics). Further details are discussed individually as follows.

\begin{table}[H]
\centering
\footnotesize{
\begin{tabular}{lcccc} 
\hline
Study             							& Language                 	& Domain   & Method        & Features \\ 
\hline
\small{Plank \& Hovy} 						& En                		& T       & lr            & w,u,n\\         
\small{ben Verhoeven et al.}    			& De,Du,It,Fr,Pt,Sp 		& T       & svm           & w,c\\  
\small{Lukito et al.} 						& In                		& T       & nb            & w,s,pos\\
\small{Alsadhan \& Skillicorn} 				& {Ar,De,Du,En,It,Fr,Pt,Sp} & T,O,E,F & svd           & w\\
\small{Gjurkovi{\'c}  \& {\v{S}}najder}    	& En                		& R       & lr,mlp,svm    & c,w,p,t,n\\ 
\small{Keh \& Cheng}   						& En                		& O       & seq           & BERT \\ 
\small{Katiyar et al.} 						& En                		& T,O     & nb            & w\\ 
\small{Wu et al.}							& En                		& R       & lr            & BERT\\
\small{Das \& Prajapati} 					& En                		& O       & ens           & w,w2v\\ 
\small{Abidin et al.} 						& En                		& O       & rf            & s\\ 
\small{Khan et al.}  						& En                		& O       & xg            & w\\ 
\small{Amirhosseini \& Kazemian}    		& En                		& O       & xg            & w \\  
\hline
\end{tabular}
}
\caption{\label{tab.related}Related work}
\end{table}

The work  in \cite{mbti-plank} is among the first of its kind to address the issue of MBTI personality classification in an open-vocabulary approach, that is, without resorting to personality lexicons or similar resources. The work addresses personality classification in the Twitter domain by using logistic regression over word n-grams, user (e.g., user's gender) and network (e.g., number of social media followers etc.) features.  Results are shown to outperform a majority class baseline.

The work  in \cite{twisty} introduces the TwiSty corpus, a large multilingual Twitter dataset labelled with MBTI information in six languages (German, Dutch, French, Italian, Portuguese, and Spanish.) The corpus conveys 34 million tweets written by over 18 thousand users. A significant portion  of the data   (about 59\%) concerns Spanish texts, which makes the other languages much less represented (e.g., 2.2\% in German, and 2.6\% in Italian.) Since some personality traits are naturally rarer than others, the corpus is also heavily imbalanced across MBTI classes. To illustrate the use of the corpus data, results from a linear SVM classifier and majority class are presented. 

The study in \cite{mbti-indon} addresses MBTI classification from Twitter data in the Indonesian language  by comparing  a number of models based on Naive Bayes classification, TF-IDF and part-of-speech counts. Among these, standard Naive Bayes text classification is found to be the overall best strategy.   

The work in \cite{Alsadhan} presents a comprehensive investigation of both Big Five and MBTI personality classification in multiple corpora and languages. The method uses single value decomposition (SVD) to discriminate extreme personality traits (e.g., introvert versus extrovert). For most languages available from the TwiSty corpus, results are found to outperform those in \cite{twisty}.

The work in \cite{mbti9k} introduces the {MBTI9K} corpus, a large collection of Reddit posts labelled with MBTI information. The corpus conveys 354.996 posts written by 9,872 users in the English language, and it is also heavily imbalanced across MBTI classes. The use of the data is illustrated by a number of experiments involving logistic regression, SVM and multi-layer perceptron (MLP) classifiers using a range of alternative text features (e.g., word and character n-grams, psycholinguistics-motivated features etc.) Results show that MLP classifiers using the entire feature set generally obtains best results.

The work in \cite{mbti-keh} is among the first to use pre-trained language models for MBTI personality classification from text, and also for personality-dependent language generation. To this end, a pre-trained BERT \cite{bert} model is fine-tuned to classify texts taken from a purpose-built dataset of online discussions about personality. The authors suggest that the BERT model presents accuracy above 70\% in the task, and point out that this is considerably superior to the results observed in other domains such as the MBTI9k Reddit  corpus \cite{mbti9k}. However, the analysis does not present any baseline results obtained from the same corpus, so it remains unclear whether the model is indeed superior to existing work, or whether  personality classification from personality-related texts (e.g., in which users presumably discuss their personality traits, personality test results etc.) may be simply more straightforward than performing the same task based on more general social media text. 

The work in \cite{mbti-stack} investigates a practical application of MBTI classification by focusing on  social media data (e.g., blogs, Twitter, and Stack Overflow) as a means to recruit project teams. To this end, a model based on Naive Bayes classification and TF-IDF counts is evaluated using a set of 40 Twitter and Stack Overflow users, whose results suggest that it may be possible to infer both personality traits and technical skills from text to facilitate recruitment.  

The work in \cite{mbti-author} addresses the issue of author-dependent word embeddings for author profiling classification by introducing a model  called {\it Author2Vec}. The possible use of this formalism is  illustrated by discussing two downstream applications, namely, depression detection and MBTI personality classification from text. The latter makes use of a logistic regression classifier built from  a subset of the MBTI9k corpus \cite{mbti9k} conveying about half of the original corpus data. The model is found to outperform a number of alternative regression models based on static Word2vec embeddings \cite{word2vec}, TF-IDF counts, and LDA topic modelling \cite{lda}.

Finally, a number of recent studies in MBTI personality classification have made use of a social media corpus available from Kaggle\footnote{{https://www.kaggle.com/datasnaek/mbti-type}} whose details regarding  domain and data collection methods remain scarce. These  include the work in \cite{mbti-ijcrt}, which compares  boosting, bagging, and stacking ensemble methods using  concatenated TF-IDF counts and word embeddings; the work in \cite{mbti-impro}, which uses random forest and features based on text statistics (e.g., sentence length, punctuation etc.); and the studies in \cite{mbti-khan} and \cite{mbti-ml},  both of which using XGBoost ensemble learning \cite{xgboost} over  word counts.

\subsection{Summary}

Existing work in MBTI personality classification from text based on pre-trained language models such as BERT remain few. The two  main exceptions are the study in \cite{mbti-keh}, which does not provide sufficiently complete evaluation details for further analysis, and the work in \cite{mbti-author}, which is mainly focused on a novel author-oriented word embedding formalism, and which uses only a small subset of the MBTI9k corpus in \cite{mbti9k} as a working example of how models of this kind may be built. This motivates a more comprehensive investigation of  the present task along these lines, and taking into account  languages other than English.

\section{Materials and methods}
\label{sec.method}

We envisaged a series of experiments in supervised machine learning from text to compare standard text classification models - based on bag-of-words and static word embeddings alike - with those built by fine-tuning a pre-trained BERT language model  to perform MBTI classification. In doing so, we would like to show that the BERT-based approach outperforms the alternatives by a large margin. 

\begin{figure}[H]
	\centering
	\includegraphics[width=0.6\textwidth]{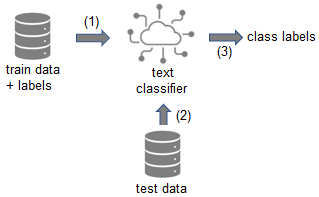}
	\caption{\label{fig.diag}Experiment pipeline.}
\end{figure}

Our experiments follow a 3-steps supervised machine learning pipeline that relies on training data (i.e., text documents) labelled with MBTI personality information to (1) build a text classifier model, (2) use the model to predict the class of previously unseen test data, and to produce (3) the corresponding output labels. This procedure is illustrated in Figure \ref{fig.diag}. 

As a means to reduce the risk of overfitting, the experiments will be carried out in a 10-fold cross-validation setting, that is, each individual text classifier is built 10 times while varying the slices of data taken as train and test sets and, at the end of the evaluation, we report mean results over the 10 execution. 

The models under discussion are to be evaluate using Reddit and Twitter text data in multiple languages. In all cases, MBTI personality detection will be modelled as a set of four independent binary classification  tasks\footnote{For instance, the EI class will be assigned the zero value when the E trait is prevalent, or the one value otherwise.} corresponding to the four MBTI personality type indicators \cite{mbti}:

\begin{enumerate}
\item {EI: Extraversion (E) versus Introversion (I)};
\item {NS: Intuition (N) versus  Sensing (S)};
\item {TF: Thinking (T) versus Feeling (F)};
\item {PJ: Perceiving (P) versus Judging (J)}.
\end{enumerate}

The following sections describe the models developed for MBTI personality classification, the  corpora to be taken as train/test data, and further details regarding the pre-processing and training procedures.

\subsection{Models}
\label{sec.models}

MBTI personality classification from text will be assessed by comparing three alternatives\footnote{Code available  from {https://github.com/vitorsantos95/mbti-classifier}}: BERT pre-trained language models, long short-term memory networks (LSTM) using static word embeddings, and a  logistic regression bag-of-words baseline. These are discussed in turn as follows.

The main focus of the present work is the use of BERT \cite{bert} pre-trained language models, which are presently fine-tuned for the MBTI personality classification task. To this end, we compute context-dependent \mbox{DistilBert} \cite{sanh2019distilbert} embeddings, and then feed 32-token input sequences  to a network conveying a 512-neuron dense layer. This is followed by a 50\% dropout layer, and by a 2-neuron dense layer that produces  the binary classification result. 

As an alternative to the use of pre-trained language models, we also consider the use of a sequence classifier based on a static word embeddings representation using LSTMs. Methods of this kind have been shown to obtain encouraging results in a wide range of NLP tasks, from stance and sentiment analysis \cite{deep-sentiment, ranlp-wesley, bracis-pavan} to author profiling \cite{b5-propor,ashraf,escobar} and others. More specifically,  we compute Word2vec \cite{word2vec} skip-gram word embeddings from each corpus\footnote{For this purpose, the corpus data is taken simply as a collection of word strings, that is, without access to any (MBTI) class information.} using a standard 300-dimension size and 8-word window. This representation is fed into a fixed LSTM architecture (i.e., with no further fine-tuning due to computational efficiency issues) comprising a 15-neuron  attention layer, two LSTM layers containing 15 neurons each,  and a 20\% dropout layer. This is  followed by a 64-neuron dense layer, and a 2-neuron softmax output layer.

Finally, we also consider  a standard bag-of-words text classifier based on logistic regression over a word n-grams text representation. This approach, hereby called \mbox{\it Reg.word}, makes use of TF-IDF n-gram  counts. Each of these input representations was subject to univariate feature selection, and the $k$ optimal features were searched in a development dataset within the 30000-1000 range at -1000 intervals using F1 as a score function. Other logistic regression parameters were kept constant by using L2  penalty, lbfgs optimisation, balanced class weights, and $10^{-4}$ tolerance.

\subsection{Data}
\label{sec.data}

The models described in the previous section are to be evaluated in multiple languages, namely, English, German, Italian, Dutch, French, Portuguese, and Spanish. In the case of English, we  use  the MBTI9k corpus of Reddit posts \cite{mbti9k}, and for the other languages we  use the TwiSty corpus in the Twitter domain \cite{twisty}. Both MBTI9k and TwiSty texts are labelled with the four MBTI personality indicators, whose class distribution is summarised in Table \ref{tab.corpus}. We notice however that both datasets are slightly smaller than those originally reported  in \cite{mbti9k} and \cite{twisty} since some of the data are no longer available online, or were removed due to noise. This issue is more prevalent in the Twitter domain in general, but it has also been raised in the context of the present Reddit dataset in \cite{mbti-author}. 

\begin{table}[H]
\centering
\setlength{\tabcolsep}{3pt} 
\begin{tabular}{ c r r  r r  r r  r r } 
\hline
{Lang.} & {E} & {I} & {N} & {S} & {T} & {F} & {P} &  {J}\\
\hline
{En} & 1,423 & 5,053 & 5,625 & 851 & 4,168 & 2,308 & 3,759 & 2,717 \\
{De} & 92,452 & 180,252 & 227,409 & 45,295 & 113,414 & 159,290 & 170,232 & 102,472 \\
{It} & 26,445 & 59,048 & 69,161 & 16,332 & 44,157 & 41,336 & 36,012 & 49,481 \\
{Du} & 70,904 & 33,589 & 76,987 & 27,506 & 35,791 & 68,702 & 66,447 & 38,046 \\
{Fr} & 249,742 & 481,480 & 566,473 & 164,749 & 297,702 & 433,520 & 451,187 & 280,035 \\
{Pt}& 27,920 & 32,387 & 44,919 & 15,388 & 27,160 & 33,147 & 32,536 & 27,771 \\
{Sp}& 243,840 & 277,788 & 334,483 & 187,145 & 199,573 & 322,055 & 302,092 & 219,536 \\
\hline
\end{tabular}
\caption{\label{tab.corpus}Corpus class distribution across English (En), German (De), Italian (it), Dutch (Du), French (Fr), Portuguese (Pt), and  Spanish (Sp) subsets.}
\end{table}

\subsection{Procedure}
\label{sec.proced}

The data from each corpus was subject to a 30/70 development/test split. Development sets were taken as an input to compute hyper-parameters for each model, and then discarded. Validation proper was performed using 10-fold cross validation over the previously unseen test sets. 

All texts were subject to the removal of special characters (in particular, emoticons). In a pilot experiment, we also found out that stop words did not generally improve results for the task at hand and, accordingly, these were removed using NLTK \cite{bird2006nltk} for the sake of efficiency. Other than that, all input texts were left unchanged. 

From the true positives (TP), true negatives (TN), false positives (FP) and false negatives (FN) obtained by each model, we computed precision, recall, F1 and accuracy scores as follows \cite{prf}.\\[2ex]

Precision = $\frac{TP}{TP+FP}$\\[2ex]

Recall = $\frac{TP}{TP+FN}$\\[2ex]

F1 = 
 $\frac{2*TP}{2*TP+FP+FN}$\\[2ex]

Accuracy = $\frac{TP+TN}{TP+TN+FP+FN}$\\[2ex]

\section{Results}
\label{sec.results}

Table \ref{tab.results} summarises results obtained by the  models discussed  in the previous sections, and also from a majority class baseline. All results were obtained by performing 10-fold cross-validation. For brevity, in what follows we only present the mean F1 scores obtained by each model. For the full results (i.e., precision, recall, F1 and accuracy scores) we report to Table \ref{tab.full.results} at the end of this article.

\begin{table}[H]
\centering
\begin{tabular}{c l c c c c c c c} 
\hline
Task & Model & En & De & Sp & Fr & It & Du & Pt\\ 
\hline
\multirow{5}{*}{EI}
& Majority  & 0.43 & 0.40 & 0.35 & 0.40 & 0.41 & 0.40 & 0.35 \\
& Reg.char  & 0.51 & 0.60 & 0.58 & 0.59 & 0.65 & 0.61 & 0.65 \\                 	
& Reg.word  & 0.54 & 0.60 & 0.58 & 0.59 & 0.63 & 0.62 & 0.64 \\
& LSTM      & 0.83 & 0.73 & 0.71 & 0.72 & 0.80 & 0.82 & 0.80 \\ 
& BERT      & \textbf{0.94} & \textbf{0.90} &  \textbf{0.86} &  \textbf{0.89} &  \textbf{0.95} &  \textbf{0.88} & \textbf{0.93}\\ 
\hline  
\multirow{5}{*}{NS}
& Majority  & 0.46 & 0.45 & 0.39 & 0.44 & 0.45 & 0.42 & 0.43\\
& Reg.char  & 0.51 & 0.58 & 0.58 & 0.56 & 0.63 & 0.60 & 0.62 \\                         
& Reg.word  & 0.54 & 0.57 & 0.58 & 0.56 & 0.64 & 0.63 & 0.61 \\ 
& LSTM      & 0.82 & 0.75 & 0.68 & 0.74 & 0.79 & \textbf{0.82} & \textbf{0.79}\\ 
& BERT      & \textbf{0.91} & \textbf{0.90} &  \textbf{0.83} &  \textbf{0.87} &  \textbf{0.89} & 0.73 & 0.75\\ 
\hline  
\multirow{5}{*}{TF}
& Majority  & 0.39 & 0.37 & 0.38 & 0.37 & 0.34 & 0.40 & 0.35 \\
& Reg.char  & 0.62 & 0.58 & 0.57 & 0.55 & 0.59 & 0.61 & 0.58 \\                        
& Reg.word  & 0.65 & 0.58 & 0.57 & 0.56 & 0.59 & 0.61 & 0.58 \\
& LSTM      & 0.82 & 0.73 & 0.69 & 0.72 & 0.78 & 0.81 & 0.81 \\ 
& BERT      & \textbf{0.89} & \textbf{0.91} &  \textbf{0.89} &  \textbf{0.88} &  \textbf{0.93} &  \textbf{0.95} &  \textbf{0.96}\\ 
\hline  
\multirow{5}{*}{PJ}  
& Majority & 0.36 & 0.38 & 0.37 & 0.38 & 0.37 & 0.39 & 0.35 \\
& Reg.char & 0.57 & 0.57 & 0.58 & 0.56 & 0.62 & 0.59 & 0.57 \\		                
& Reg.word & 0.60 & 0.58 & 0.57 & 0.56 & 0.61 & 0.60 & 0.58 \\
& LSTM     & 0.82 & 0.72 & 0.69 & 0.70 & 0.78 & 0.80 & 0.79 \\ 
& BERT     & \textbf{0.91} & \textbf{0.86} & \textbf{0.83} &  \textbf{0.74} &  \textbf{0.93} &  \textbf{0.91} & \textbf{0.94}\\ 
\hline  
\end{tabular}
\caption{\label{tab.results}10-fold cross-validation mean F1 results for English (En), German (De), Spanish (Sp), French (Fr), Italian (It), Dutch (Du), and Portuguese (Pt) data. Best F1 results for each class and language are highlighted.}
\end{table}

Results from Table \ref{tab.results} show that BERT generally outperforms the alternatives in all but two cases - the NS task in Dutch (Du) and Portuguese (Pt) - in which case the LSTM model obtained overall best results. According to a McNemar test \cite{mcnemar}, all differences between BERT and LSTM are statistically significant at $p < 0.001$ except for the NS task in Italian (It), in which case the difference is significant at  $p < 0.005$ only. This outcome offers support to our main research question, that is, the use of pre-trained language models for MBTI personality classification outperforms standard text classification methods based on bag-of-words and static word embeddings alike.

As  discussed in Section \ref{sec.method}, the present experiments could not use exactly the same datasets as in previous work due to the  removal of social media texts over time and, as a result, a direct comparison is not entirely possible. Bearing this limitation in mind, Table \ref{tab.comp.en} presents - purely for illustration purposes - mean F1 results reported in \cite{mbti9k} and \cite{mbti-author} for the English MBTI9k corpus alongside those obtained by our present BERT model. Similarly, Table \ref{tab.comp.tw} presents weighted F1 results reported in both \cite{twisty} and \cite{Alsadhan} for the TwiSty corpus alongside present BERT.

\setlength{\tabcolsep}{8pt} 
\begin{table}[H]
\centering
\begin{tabular}{c c c c}
\hline
Task 	& Gjurkovi{\'c}  \& {\v{S}}najder 	& Wu et. al. & Current (BERT)\\ 
\hline
EI 		& 0.83 			&  0.69        & 0.94 \\
NS 		& 0.79 			&  0.77        & 0.91 \\
TF 		& 0.64 			&  0.68        & 0.89 \\
PJ 		& 0.74 			&  0.61        & 0.91 \\ 
\hline
\end{tabular}
\caption{\label{tab.comp.en}MBTI9k mean F1 results from previous work \cite{mbti9k} and \cite{mbti-author}, and from the present BERT models.}
\end{table}

\setlength{\tabcolsep}{8pt} 
\begin{table}[H]
\centering
\begin{tabular}{c c c c c}
\hline
Lang. & Task & Verhoeven& Alsadhan  & Current (BERT)\\ 
\hline
\multirow{4}{*}{De} 	& EI & 0.72 & 0.76 & 0.77 \\
                    	& NS & 0.74 & 0.78 & 0.93 \\
                    	& TF & 0.59 & 0.78 & 0.87 \\
                    	& PJ & 0.62 & 0.80 & 0.92 \\ 
\hline
\multirow{4}{*}{Sp} 	& EI & 0.61 & 0.72 & 0.84 \\
                      	& NS & 0.62 & 0.73 & 0.91 \\
                      	& TF & 0.60 & 0.72 & 0.79 \\
                      	& PJ & 0.56 & 0.69 & 0.88 \\ 
\hline
\multirow{4}{*}{Fr} 	& EI & 0.66 & 0.86 & 0.78 \\
                     	& NS & 0.79 & 0.96 & 0.92 \\
                     	& TF & 0.58 & 0.74 & 0.81 \\
                     	& PJ & 0.57 & 0.84 & 0.86 \\ 
\hline
\multirow{4}{*}{It} 	& EI & 0.78 & 0.90 & 0.88 \\
                      	& NS & 0.79 & 0.67 & 0.95 \\
                      	& TF & 0.52 & 0.83 & 0.93 \\
                      	& PJ & 0.47 & 0.79 & 0.91 \\ 
\hline
\multirow{4}{*}{Du} 	& EI & 0.63 & 0.85 & 0.94 \\
                      	& NS & 0.70 & 0.94 & 0.97 \\
                      	& TF & 0.60 & 0.82 & 0.91 \\
                      	& PJ & 0.58 & 0.87 & 0.94 \\ 
\hline
\multirow{4}{*}{Pt} 	& EI & 0.67 & 0.85 & 0.92 \\
                      	& NS & 0.73 & 0.94 & 0.93 \\
                      	& TF & 0.62 & 0.80 & 0.94 \\
                      	& PJ & 0.57 & 0.88 & 0.95 \\ 
\hline
\end{tabular}
\caption{\label{tab.comp.tw}TwiSty weighted F1 results from previous work  \cite{twisty} and \cite{Alsadhan},  and from the present BERT models for the German (De), Spanish (Sp), French (Fr), Italian (It), Dutch (Du), and Portuguese (Pt) languages.}
\end{table}

As a means to illustrate the most relevant word features for each task, we performed eli5 prediction explanation\footnote{{https://eli5.readthedocs.io/en/latest/}} to compute the terms more strongly correlated with each class, using as an example the word-based  \mbox{\it Reg.word} classifier and the English \mbox{MBTI9k} dataset. This, despite being outperformed by our main BERT models, is more suitable to interpretation. 

Selected features are illustrated in Table \ref{tab.eli5}, in which  word weights represent the change (decrease/increase) of the evaluation score when the specific feature is shuffled, keeping in mind that MBTI classes are not independent and, due to class imbalance, words that would intuitively be more associated with a particular MBTI type may have been selected by association with another, concomitant type (e.g., if users labelled as extraverts also happen to be mostly of the thinking type etc.)

\setlength{\tabcolsep}{4pt} 
\begin{table}[H]
\centering
\begin{tabular}{c l | c l | cl | cl}
\hline
Weight & EI & Weight & NS & Weight & TF & Weight & PJ \\
\hline
+0.113 & job & +0.051 & yourself & +0.035 & thank & +0.051 & comcast \\
+0.097 & may & +0.047 & after & +0.033 & baby & +0.036 & 25b2 \\
+0.090 & never & +0.047 & job & +0.030 & amazing & +0.034 & now \\
+0.086 & them & +0.046 & trans & +0.027 & still & +0.033 & nice \\
+0.084 & free & +0.046 & up & +0.025 & two & +0.033 & story \\
+0.081 & im & +0.046 & etc & +0.025 & pregnancy & +0.032 & awesome \\
+0.080 & aren & +0.045 & than & +0.025 & actually & +0.032 & always \\
+0.076 & 10 & +0.044 & all & +0.024 & team & +0.029 & others \\
+0.073 & couple & +0.043 & end & +0.024 & from & +0.029 & isn \\
+0.073 & wiki & +0.041 & know & +0.023 & these & +0.029 & game \\
...    & ...     & ... & ... & ... & ... & ... & ... \\
-0.058 & comcast & -0.030 & honestly & -0.018 & how & -0.022 & let \\
-0.058 & temple & -0.030 & without & -0.018 & fucking & -0.022 & well \\
-0.059 & death & -0.030 & run & -0.018 & ve & -0.022 & nothing \\
-0.061 & est & -0.030 & help & -0.018 & lol & -0.023 & soylent \\
-0.062 & city & -0.031 & amazing & -0.018 & point & -0.023 & edit \\
-0.062 & vs & -0.031 & running & -0.019 & something & -0.023 & still \\
-0.062 & week & -0.031 & totally & -0.019 & mind & -0.023 & sex \\
-0.062 & usually & -0.032 & thing & -0.019 & op & -0.023 & completely \\
-0.062 & own & -0.032 & won & -0.020 & lt & -0.023 & lmao \\
-0.062 & album & -0.032 & which & -0.020 & female & -0.024 & clinton \\
\hline
\end{tabular}
\caption{\label{tab.eli5}Top-10 positive and negative word weights for each classification task using the {\it Reg.word}  logistic regression classifier.}
\end{table}

%
%
%

Finally, the following text examples illustrate correctly classified instances for each MBTI type, in which the most relevant (word) features are highlighted. 

\begin{figure}[H]
\centering
\includegraphics[width=0.9\textwidth]{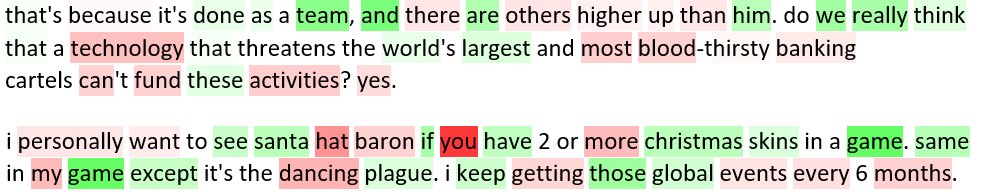}
\caption{\label{fig.ei}Extraversion (top) and introversion (bottom) features.}
\end{figure}

In Figure \ref{fig.ei} Extraversion correlates positively with 'team', and negatively with 'technology', whereas Introversion correlates positively with `game' and negatively with `dancing'. 
	
\begin{figure}[H]
	\centering
	\includegraphics[width=0.9\textwidth]{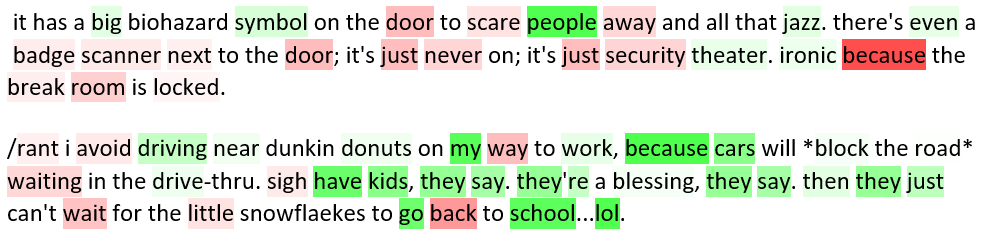}
	\caption{\label{fig.ns}Intuition (top) and Sensing (bottom) features.}
\end{figure}

In Figure \ref{fig.ns}, Intuition correlates positively with  `people' and negatively with `because' (which suggests reasoning), whereas Sensing strongly correlates with `because' and more concrete concepts (e.g., cars, school, kids etc.) 	
	
\begin{figure}[H]
	\centering
	\includegraphics[width=0.9\textwidth]{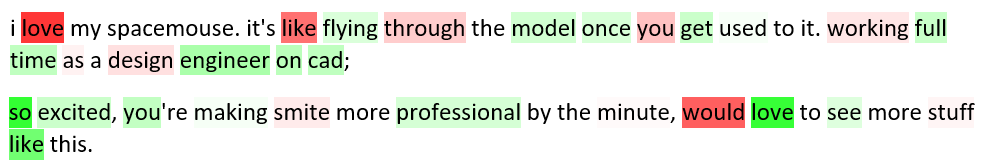}
	\caption{\label{fig.tf}Thinking (top) and Feeling (bottom) features.}
\end{figure}

In Figure \ref{fig.tf}, Thinking correlates positively with concrete concepts (e.g., model, engineer) and negatively with sentiment-charged words (e.g., `like', `love'.) By contrast, Feeling correlates positively with sentiment, and negatively with `would' (which might suggest reasoning.)  

\begin{figure}[H]
	\centering
	\includegraphics[width=0.9\textwidth]{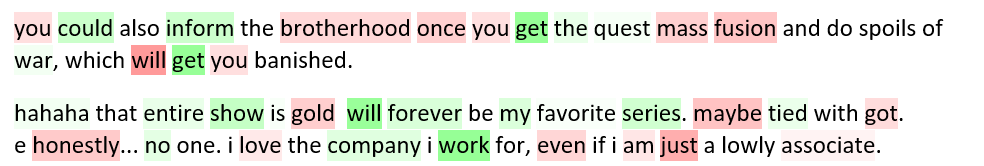}
	\caption{\label{fig.pj}Perceiving (top) and Judging (bottom) features.}
\end{figure}

Finally, in Figure \ref{fig.pj}, Perceiving correlates with a certain preference to take in information (e.g., `inform'), and Judging correlates positively with decision-making terms (e.g., `work'). 

These examples, although suboptimal for the reasons discussed above, in our view suggest a reasonable consistency with the MBTI guidelines. Examples in which the logistic regression classifier does not make the right decision, by contrast, include, the selection of 'party' as a prominent feature for Extraversion even when the term is used in its political (and not leisure) sense.  In the case of our BERT model, however, errors of this kind are arguably less likely to occur given the model's context sensitivity.

\section{Final remarks}
\label{sec.final}

This work has addressed the issue of MBTI personality classification from text with the aid of pre-trained BERT language models. The present approach has been compared against alternatives based on bag-of-words and static word embeddings representations, and its results were found to be consistently superior in a number of evaluation scenarios involving multiple target languages in the Reddit and Twitter domains, and by a significant margin.

Despite the positive initial results, however, the current set of experiments is  only a first step towards more general, domain-independent MBTI personality classification from text. One obvious limitation of the  present approach is, for instance, the focus on only two MBTI language resources (namely, the MBTI9k \cite{mbti9k} and TwiSty \cite{twisty} corpora.) Although covering seven languages in two linguistic domains, more work needs be done to investigate the present task in other text genres. 

Furthermore, we notice that the present discussion has  been limited to the issue of fine-tuning BERT models for the MBTI classification task for which training data is readily available. Outside the present Twitter and Reddit domains, however, text corpora labelled with MBTI information may be scarce, and it may be necessary to resort do domain adaptation methods. These may include, for instance, the use of BERT-based adversarial adaptation with distillation (AAD) method proposed in  \cite{BERTADA} for cross-domain sentiment analysis, among others. A study along these lines in the context of the present personality classification task is also left as future work.

Finally, we notice that in recent years there has been a surge in transformer-based language models, including ELMo \cite{elmo}, XLNet \cite{xlnet}, RoBERTa \cite{roberta}, GPT-3 \cite{gpt3}, and many others. In most cases, these models are yet to be applied to the present MBTI classification task.

\section{Acknowledgements}
This work has received support from the University of S\~ao Paulo.


\begin{landscape}

\setlength{\tabcolsep}{2pt}
\begin{table}[p]
	\centering
	\scalebox{0.7}{
	\begin{tabular}{c l | c c c c | c c c c | c c c c | c c c c | c c c c | c c c c | c c c c } 
		\hline
		\multicolumn{2}{c}{}&
		\multicolumn{4}{|c}{English}&
		\multicolumn{4}{|c}{German}&
		\multicolumn{4}{|c}{Spanish}&
		\multicolumn{4}{|c}{French}&
		\multicolumn{4}{|c}{Italian}&
		\multicolumn{4}{|c}{Dutch}&
		\multicolumn{4}{|c}{Portuguese}\\
		Task & Model & Acc & P & R & F1 & Acc & P & R & F1 & Acc & P & R & F1 & Acc & P & R & F1 & Acc & P & R & F1 & Acc & P & R & F1 & Acc & P & R & F1\\ 
		\hline
		\multirow{5}{*}{EI}
		& Majority  & 0.77 & 0.77 & 1.00 & 0.43 & 0.66 & 0.66 & 1.00 & 0.40 & 0.53 & 0.53 & 1.00 & 0.35 & 0.65 & 0.65 & 1.00 & 0.40 & 0.69 & 0.69 & 1.00 & 0.41 & 0.67 & 0.67 & 1.00 & 0.40 & 0.53 & 0.53 & 1.00 & 0.35 \\
		& Reg.char  & 0.56 & 0.55 & 0.26 & 0.51 & 0.63 & 0.49 & 0.46 & 0.60 & 0.59 & 0.57 & 0.56 & 0.58 & 0.63 & 0.46 & 0.46 & 0.59 & 0.70 & 0.42 & 0.52 & 0.65 & 0.68 & 0.80 & 0.75 & 0.61 & 0.65 & 0.58 & 0.63 & 0.65 \\                 	
		& Reg.word  & 0.54 & 0.58 & 0.26 & 0.54 & 0.63 & 0.52 & 0.46 & 0.60 & 0.58 & 0.61 & 0.54 & 0.58 & 0.64 & 0.37 & 0.46 & 0.59 & 0.71 & 0.47 & 0.55 & 0.63 & 0.68 & 0.82 & 0.73 & 0.62 & 0.66 & 0.57 & 0.65 & 0.64 \\
		& LSTM      & 0.99 & 0.98 & 0.98 & 0.83 & 0.80 & 0.56 & 0.79 & 0.73 & 0.72 & 0.66 & 0.72 & 0.71 & 0.96 & 0.48 & 0.80 & 0.72 & 0.88 & 0.72 & 0.87 & 0.80 & 0.88 & 0.96 & 0.88 & 0.82 & 0.88 & 0.86 & 0.87 & 0.80 \\ 
		& BERT      & 0.87 & 0.67 & 0.69 & \textbf{0.94} & 0.86 & 0.69 & 0.86 & \textbf{0.90} & 0.85 & 0.82 & 0.85 & \textbf{0.86} & 0.86 & 0.72 & 0.84 & \textbf{0.89} & 0.93 & 0.84 & 0.93 & \textbf{0.95} & 0.92 & 0.96 & 0.93 & \textbf{0.88} & 0.92 & 0.91 & 0.92 & \textbf{0.93}\\ 
		\hline  
		\multirow{5}{*}{NS}
		& Majority  & 0.86 & 0.86 & 0.86 & 0.46  & 0.83 & 0.83 & 0.83 & 0.45  & 0.64 & 0.64 & 0.64 & 0.39  & 0.77 & 0.77 & 0.77 & 0.44  & 0.80 & 0.80 & 0.80 & 0.45 & 0.73 & 0.73 & 0.73 & 0.42 & 0.74 & 0.74 & 0.74 & 0.43\\
		& Reg.char  & 0.63 & 0.64 & 0.90 & 0.51 & 0.75 & 0.84 & 0.86 & 0.58 & 0.61 & 0.66 & 0.71 & 0.58 & 0.74 & 0.89 & 0.79 & 0.56 & 0.81 & 0.93 & 0.85 & 0.63 & 0.71 & 0.80 & 0.80 & 0.60 & 0.68 & 0.73 & 0.81 & 0.62 \\                         
		& Reg.word  & 0.66 & 0.70 & 0.89 & 0.54 & 0.78 & 0.89 & 0.85 & 0.57 & 0.61 & 0.69 & 0.69 & 0.58 & 0.73 & 0.87 & 0.80 & 0.56 & 0.76 & 0.84 & 0.86 & 0.64 & 0.70 & 0.81 & 0.79 & 0.63 & 0.68 & 0.72 & 0.82 & 0.61 \\ 
		& LSTM      & 0.99 & 0.99 & 0.99 & 0.82 & 0.85 & 0.90 & 0.85 & 0.75 & 0.76 & 0.93 & 0.75 & 0.68 & 0.85 & 0.97 & 0.85 & 0.74 & 0.92 & 0.96 & 0.93 & 0.79 & 0.90 & 0.95 & 0.91 & \textbf{0.82} & 0.91 & 0.96 & 0.92 & \textbf{0.79}\\ 
		& BERT      & 0.95 & 0.97 & 0.95 & \textbf{0.91} & 0.92 & 0.94 & 0.93 & \textbf{0.90} & 0.88 & 0.93 & 0.89 & \textbf{0.83} & 0.90 & 0.93 & 0.90 & \textbf{0.87} & 0.93 & 0.98 & 0.92 & \textbf{0.89} & 0.96 & 0.96 & 0.98 & 0.73 & 0.89 & 0.97 & 0.89 & 0.75\\ 
		\hline  
		\multirow{5}{*}{TF}
		& Majority  & 0.64 & 0.64 & 0.64 & 0.39 & 0.58 & 0.66 & 0.66 & 0.37 & 0.61 & 0.53 & 0.53 & 0.38  & 0.59 & 0.59 & 0.59 & 0.37 & 0.51 & 0.51 & 0.51 & 0.34 & 0.65 & 0.65 & 0.65 & 0.40 & 0.54 & 0.54 & 0.54 & 0.35 \\
		& Reg.char  & 0.64 & 0.66 & 0.75 & 0.62 & 0.59 & 0.61 & 0.50 & 0.58 & 0.59 & 0.49 & 0.47 & 0.57 & 0.59 & 0.41 & 0.50 & 0.55 & 0.60 & 0.73 & 0.59 & 0.59 & 0.65 & 0.49 & 0.49 & 0.61 & 0.64 & 0.68 & 0.58 & 0.58 \\                        
		& Reg.word  & 0.65 & 0.65 & 0.76 & 0.65 & 0.58 & 0.58 & 0.50 & 0.58 & 0.60 & 0.45 & 0.47 & 0.57 & 0.58 & 0.38 & 0.48 & 0.56 & 0.61 & 0.71 & 0.60 & 0.59 & 0.66 & 0.48 & 0.50 & 0.61 & 0.64 & 0.68 & 0.59 & 0.58 \\
		& LSTM      & 0.99 & 0.99 & 0.99 & 0.82 & 0.77 & 0.63 & 0.77 & 0.73 & 0.75 & 0.47 & 0.80 & 0.69 & 0.79 & 0.51 & 0.97 & 0.72 & 0.84 & 0.85 & 0.84 & 0.78 & 0.87 & 0.76 & 0.85 & 0.81 & 0.88 & 0.86 & 0.88 & 0.81 \\ 
		& BERT      & 0.93 & 0.95 & 0.94 & \textbf{0.89} & 0.89 & 0.86 & 0.88 & \textbf{0.91} & 0.85 & 0.75 & 0.83 & \textbf{0.89} & 0.85 & 0.78 & 0.85 & \textbf{0.88} & 0.93 & 0.94 & 0.93 & \textbf{0.93} & 0.94 & 0.89 & 0.92 & \textbf{0.95} & 0.95 & 0.94 & 0.94 & \textbf{0.96}\\ 
		\hline  
		\multirow{5}{*}{PJ}  
		& Majority  & 0.58 & 0.58 & 0.58 & 0.36 & 0.62 & 0.62 & 0.62 & 0.38 & 0.57 & 0.57 & 0.57 & 0.37 & 0.61 & 0.61 & 0.61 & 0.38 & 0.57 & 0.57 & 0.57 & 0.37 & 0.63 & 0.63 & 0.63 & 0.39 & 0.53 & 0.53 & 0.53 & 0.35 \\
		& Reg.char  & 0.57 & 0.54 & 0.65 & 0.57 & 0.60 & 0.65 & 0.69 & 0.57 & 0.59 & 0.66 & 0.64 & 0.58 & 0.61 & 0.76 & 0.66 & 0.56 & 0.62 & 0.57 & 0.55 & 0.62 & 0.65 & 0.80 & 0.69 & 0.59 & 0.58 & 0.64 & 0.61 & 0.57 \\                        
		& Reg.word  & 0.59 & 0.61 & 0.66 & 0.60 & 0.61 & 0.73 & 0.67 & 0.58 & 0.59 & 0.65 & 0.64 & 0.57 & 0.60 & 0.74 & 0.65 & 0.56 & 0.63 & 0.49 & 0.58 & 0.61 & 0.64 & 0.80 & 0.69 & 0.60 & 0.60 & 0.66 & 0.62 & 0.58 \\
		& LSTM      & 0.99 & 0.99 & 0.99 & 0.82 & 0.77 & 0.92 & 0.76 & 0.73 & 0.73 & 0.90 & 0.71 & 0.69 & 0.76 & 0.93 & 0.74 & 0.72 & 0.84 & 0.78 & 0.84 & 0.78 & 0.86 & 0.93 & 0.86 & 0.81 & 0.87 & 0.90 & 0.87 & 0.81 \\ 
		& BERT      & 0.91 & 0.92 & 0.93 & \textbf{0.91} & 0.90 & 0.93 & 0.90 & \textbf{0.86} & 0.86 & 0.90 & 0.86 & \textbf{0.83} & 0.82 & 0.91 & 0.82 & \textbf{0.74} & 0.93 & 0.91 & 0.92 & \textbf{0.93} & 0.93 & 0.95 & 0.94 & \textbf{0.91} & 0.94 & 0.95 & 0.94 & \textbf{0.94}\\ 
		\hline  
	\end{tabular}
}
	\caption{\label{tab.full.results}10-fold cross-validation mean accuracy (Acc), precision (P), recall (R) and F1 results. Best F1 results for each class and language are highlighted.}
\end{table}

\end{landscape}

\clearpage


\begingroup

\endgroup

\end{document}